\documentclass[default]{sn-jnl}

\usepackage{graphicx}%
\usepackage{multirow}%
\usepackage{amsmath,amssymb,amsfonts}%
\usepackage{amsthm}%
\usepackage{mathrsfs}%
\usepackage[title]{appendix}%
\usepackage{xcolor}%
\usepackage{textcomp}%
\usepackage{manyfoot}%
\usepackage{booktabs}%
\usepackage{algorithm}%
\usepackage{algorithmicx}%
\usepackage{algpseudocode}%
\usepackage{listings}%
\usepackage{multicol}
\usepackage{amsmath}
\usepackage{wasysym}
\usepackage{utfsym}

\theoremstyle{thmstyleone}%
\theoremstyle{thmstyletwo}%

\theoremstyle{thmstylethree}%

\begin{document}

\title[Article Title]{Dual-View Pyramid Pooling in Deep Neural
Networks for Improved Medical Image Classification and Confidence Calibration}


\author[1,2]{\fnm{Xiaoqing} \sur{Zhang}}
\equalcont{These authors contributed equally to this work.}

\author[2]{\fnm{Qiushi} \sur{Nie}}
\equalcont{These authors contributed equally to this work.}

\author[2]{\fnm{Zunjie} \sur{Xiao}}

\author[2]{\fnm{Jilu} \sur{Zhao}}
\author[2]{\fnm{Xiao} \sur{Wu}}
\author[3]{\fnm{Pengxin} \sur{Guo}}

\author[4]{\fnm{Runzhi} \sur{Li}}
\author[5,6]{\fnm{Jin} \sur{Liu}}
\author*[1]{\fnm{Yanjie} \sur{Wei}}\email{yj.wei@siat.ac.cn}

\author*[1,7]{\fnm{Yi} \sur{Pan}}\email{yi.pan@siat.ac.cn}

\affil[1]{\orgdiv{Center for High Performance Computing and Shenzhen Key Laboratory of Intelligent Bioinformatics, 
}, \orgname{Shenzhen Institute of Advanced Technology, Chinese Academy of Sciences}, \orgaddress{\city{Shenzhen}, \postcode{518055}, \state{Guangdong}, \country{China}}}

\affil[2]{\orgdiv{Department of Computer Science and Engineering}, \orgname{Southern University of Science and Technology}, \orgaddress{\city{Shenzhen}, \postcode{518055},  \country{China}}}

\affil[3]{\orgdiv{Department of Statistics and Actuarial Science}, \orgname{The University of Hong Kong}, \orgaddress{\city{Hong Kong}, \postcode{ 999077},  \country{China}} }

\affil[4]{\orgdiv{Cooperative Innovation Center of Internet Healthcare}, \orgname{Zhengzhou University}, \orgaddress{\city{Zhengzhou}, \postcode{450001}, \country{China}}}

\affil[5]{\orgdiv{Hunan Provincial Key Lab on Bioinformatics, School of Computer Science and Engineering}, \orgname{Central South University}, \orgaddress{\city{Changsha}, \postcode{410083}, \country{China}}}

\affil[6]{\orgdiv{Xinjiang Engineering Research Center of Big Data  and Intelligent Software, School of software}, \orgname{Xinjiang University}, \orgaddress{\city{Wulumuqi}, \postcode{830046}, \country{China}}}

\affil[7]{\orgdiv{Faculty of Computer Science and Control Engineering}, \orgname{Shenzhen University of Advanced Technology}, \orgaddress{\city{Shenzhen}, \country{China}}}

\abstract{Spatial pooling (SP) and cross-channel pooling (CCP) operators have been applied to aggregate spatial features and pixel-wise features from feature maps in deep neural networks (DNNs), respectively. Their main goal is to reduce computation and memory overhead without visibly weakening the performance of DNNs. However, SP often faces the problem of losing the subtle feature representations, while CCP has a high possibility of ignoring salient feature representations, which may lead to both miscalibration of confidence issues and suboptimal medical classification results. To address these problems, we propose a novel dual-view framework, the first to systematically investigate the relative roles of SP and CCP by analyzing the difference between spatial features and pixel-wise features. Based on this framework, we propose a new pooling method, termed dual-view pyramid pooling (DVPP), to aggregate multi-scale dual-view features. DVPP aims to boost both medical image classification and confidence calibration performance by fully leveraging the merits of SP and CCP operators from a dual-axis perspective.
Additionally, we discuss how to fulfill DVPP with five parameter-free implementations. Extensive experiments on six 2D/3D medical image classification tasks show that our DVPP surpasses state-of-the-art pooling methods in terms of medical image classification results and confidence calibration across different DNNs.
}





\maketitle

\section{Introduction}
\label{sec1}

Over the past years, deep neural networks (DNNs), such as convolutional neural networks (CNNs) \cite{9451544,zhang2022attention,zhang2024regional}, vision transformers (ViTs)~\cite{he2023transformers,shen2023structure}, and recent vision foundation models~\cite{yang2023aim,zhou2023foundation}, have achieved significant success in various applications, including visual classification and medical image classification tasks ~\cite{zhang2022attention, song2021deep,shamshad2023transformers,wiliem2014automatic}.

For the classification tasks, it is a common routine to utilize spatial pooling (SP) operators to squeeze/compress 2D/3D high-level feature map representations of DNNs into 1D spatial features, which are then fed directly into the classifier ~\cite{nirthika2022pooling}. This is mainly because, compared with fully connected (FC) layers, SP operators usually achieve competitive performance with less complexity. Global average pooling (GAP) is one of the most widely used SP operators, which aggregates the global average spatial features by compressing high-level feature map representations across spatial dimensions. Following GAP, several improved SP variants have been developed to improve the classification performance \cite{he2015spatial, 7927440, stergiou2021refining, LIP_2019_ICCV}. For example, mixed pooling (MP) \cite{7927440} extracts global average and maximum spatial features from high-level feature map representations by taking advantage of GAP and global max pooling (GMP). Spatial pyramid pooling (SPP) \cite{he2015spatial} utilizes multi-level spatial pooling operations to extract multi-scale spatial feature representations by considering the relative importance of local and global spatial features. Although spatial features aggregated by different SP operators contain salient representations, the inherent lossy properties of SP inhibit subtle feature representation preservation.

Cross-channel pooling (CCP) is another pooling paradigm that can aggregate pixel-wise features from feature maps ~\cite{chen2019condensation}. Cross-channel average pooling (CAP) operator, a representative implementation of CCP, has been utilized to aggregate pixel-wise average features across channels in the efficient CNN architecture design. The convolutional bottleneck attention module (CBAM) \cite{woo2018cbam} utilizes CAP and cross-channel maximum pooling operators to extract pixel-wise average and maximum features from feature maps to highlight informative pixel positions. Additionally, the pixel-wise features obtained by the CCP operator contain rich subtle feature representations, but they often lack the ability to highlight salient feature representations.


\begin{figure}[!t]
\centering
\includegraphics[width=0.9\textwidth,height=4.8cm]{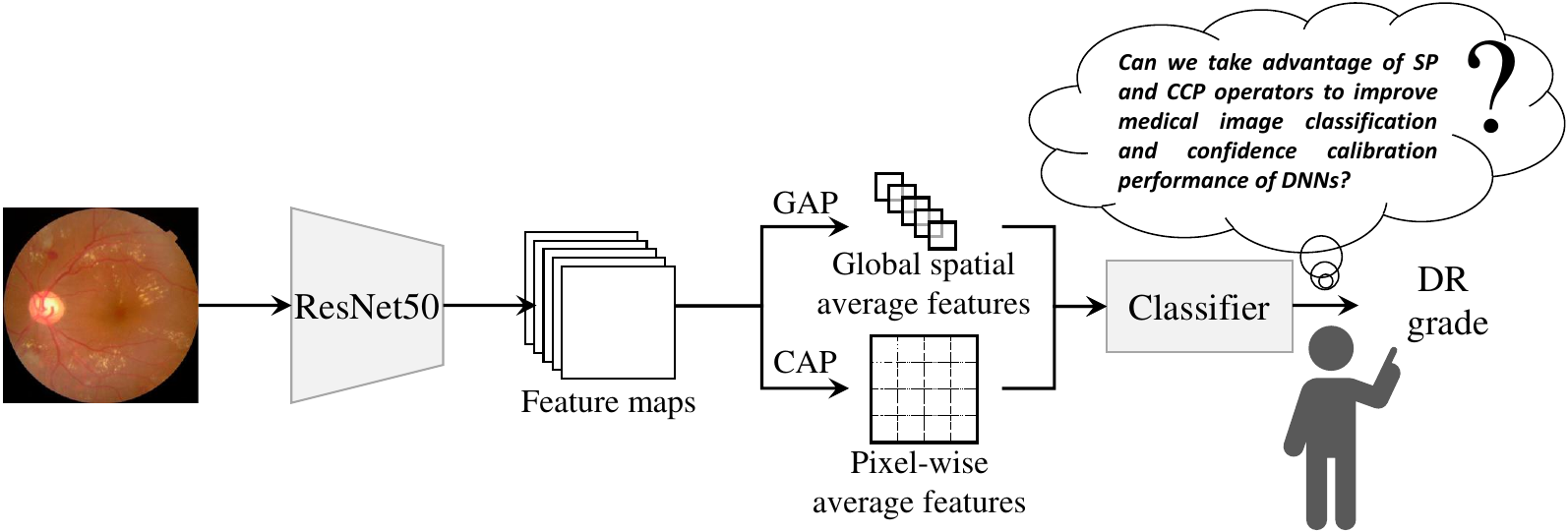}
\caption{Given a fundus image as the input, we apply GAP and CAP operators to aggregate global spatial average features and pixel-wise average features from the high-level feature maps of the last convolutional layer, including salient and subtle feature representations respectively. Here, we take the pre-trained ResNet50 as the backbone architecture and tackle the diabetic retinopathy (DR) grading task on the APTOS2019 dataset.}
\label{fig1}
\end{figure}

According to the above analysis, we gain the following insights:
\textbf{\textit{\romannumeral1)} Complementary Feature Representation Types.} SP and CCP operators complement each other regarding the feature representations they extract. To illustrate their complementarity, Fig.~\ref{fig1} visually explains the difference between global average spatial features and pixel-wise average features aggregated by GAP and CAP operators, respectively. The global average spatial feature distribution is one-dimensional (1D), including salient feature representations, whereas the pixel-wise average feature distribution is two-dimensional (2D), including subtle feature representations. This demonstrates that their feature representation types are different, each with unique significance.
\textbf{\textit{\romannumeral2)} Significance of Subtle and Salient Lesions in Medical Diagnosis.} In medical diagnosis, both subtle and salient lesions play crucial roles in assisting clinicians in making accurate and trustworthy diagnoses. These lesions can be viewed as different forms of subtle and salient feature representations.
\textbf{\textit{\romannumeral3)} Confidence Calibration Issues in Modern DNNs.} Modern DNNs often exhibit poor calibration in their predictions \cite{guo2017calibration, zhong2021mislas, patra2023calibrating} is a significant problem in computer-aided medical decision-making systems. One primary reason is that these DNNs typically use the SP operator to aggregate spatial features from high-level feature maps, inevitably overlooking pixel-wise feature extraction. Therefore, we suggest that SP and CCP operators together may impact medical classification and confidence calibration results of DNNs from the aspect of representation learning.

\emph{\textbf{Surprisingly, according to the extensive literature review, we found that no previous works have exploited SP and CCP operators simultaneously to boost medical image classification performance, not to mention the confidence calibration improvement of DNNs. We ponder: Can we design a new pooling method by integrating merits of SP and CCP operators to improve medical image classification and confidence calibration performance of DNNs?}}

In seeking answers to these problems, we propose a unique yet simple dual-view framework, which is the first time to rethink the difference between SP and CCP operators by analyzing spatial features and pixel-wise features systematically. Under this framework, this paper proposes a new yet effective pooling method, termed dual-view pyramid pooling (DVPP), to aggregate multi-scale dual-view features from high-level feature maps through a dual-axis perspective. Our method aims to improve medical image classification results and confidence calibration by fully leveraging the potential of SP and CCP operators. Note that multi-scale dual-view features involve both multi-scale spatial features and multi-scale pixel-wise features together.
Moreover, efficiently aggregating multi-scale dual-view features in the DVPP design is challenging. In this way, two naïve design paradigms of DVPP are considered: parameter-free and parametric. Unfortunately, the parametric DVPP design does not work as well as we expect in experiments, so we mainly focus on parameter-free DVPP design in this paper. Moreover, we developed five types of parameter-free DVPP designs to better mine multi-scale dual-view features together from the dual-axis perspective. Our DVPP method can be directly integrated at the end of various DNNs and trained end-to-end. Extensive experiments on six publicly available 2D/3D medical image benchmarks demonstrate that our proposed DVPP significantly boosts medical image classification results and confidence calibration compared to state-of-the-art (SOTA) pooling methods and calibration methods, keeping consistent with our expectations.

To summarize, the main contributions of this paper are listed as follows:
\begin{itemize}
  
\item This paper is the first to provide a unique dual-view framework to rethink SP and CCP operators by figuring out the difference between spatial and pixel-wise features. We argue that these two pooling operators affect medical image classification results and confidence calibration from the representation learning aspect, a \textit{research direction} has not been previously studied.

\item We propose a new yet effective pooling method, DVPP, effectively exploring multi-scale dual-view features from a dual-axis perspective to improve classification e and confidence calibration performance. Additionally, five kinds of parameter-free implementations are developed to fulfill our proposed DVPP.

\item The extensive experiments on six publicly available 2D/3D medical image benchmarks demonstrate the superiority of our proposed DVPP over SOTA pooling methods and calibration methods regarding medical image classification and confidence calibration performance. Moreover, our proposed method offers a new solution to improve both the classification and confidence calibration of DNNs, prompting the development of the confidence calibration field. We also provide a comprehensive visual analysis to explain the internal behavior of our method.

\end{itemize}

The rest of this paper is organized as follows. Section~\ref{sec2} briefly reviews related works of pooling and DNN calibration. Section~\ref{sec3} introduces our proposed DVPP in detail. Datasets, experiment settings, experimental results and analysis are presented in Section~\ref{sec:exper}. We conclude this paper in Section~\ref{sec:con}.

\section{Related Works}
\label{sec2}

\subsection{Pooling}
The existing pooling methods used in DNNs can roughly be grouped into two paradigms: SP and CCP. 

\textbf{Spatial Pooling.} 
SP operators have been typically utilized to down-sample feature maps \cite{zafar2022comparison, nirthika2022pooling} for reducing computation and memory overhead in DNN architecture construction. GAP and GMP are widely used SP methods for aggregating global spatial features from high-level feature maps in DNNs. Following them, Matthew et al.~\cite{sp} proposed stochastic pooling to randomly select the activation in each pooling region based on multinomial distribution. Kobayashi~\cite{Kobayashi_2019_ICCV} presented a local pooling operator to obtain global feature statistics for adjusting the learnable parameters for highlighting significant spatial features. Zhai~\cite{8099909} proposed stochastic spatial sampling (S3Pool) to downsample feature maps through a non-uniform sampling method. Graham \cite{graham2014fractional} proposed fractional pooling by using mixed pooling strides of 1 and 2 at different pixel locations. Stergiou et al.~\cite{stergiou2022adapool} proposed an AdaPool to retain significant details and the local structure information. Hyun et al. \cite{hyun2021universal} proposed a universal pooling (UPool) by applying a local spatial attention module to select informative features dynamically. Zhao et al. \cite{zhao2021liftpool} proposed the LiftPool by decomposing a feature map into four learnable sub-bands.

\textbf{Cross-channel Pooling.} The CCP operator has been applied to extract pixel-wise features from feature maps, which is mainly used for spatial attention block design. CBAM~ \cite{woo2018cbam} aggregated pixel-wise average and maximum features from feature maps with two different CCP operators for emphasizing significant pixel locations. Wang et al. \cite{9229188} used a CAP method to aggregate pixel-wise average features in the parameter-free spatial attention block. Zhao et al. ~\cite{zhao2023clinical} also proposed a clinical cross-channel pooling method to obtain different pixel-wise feature types for spatial attention design. Zhang et al.~\cite{Zhang2023ppcr} developed a cross-channel pyramid pooling (CCPP) method to extract multi-scale pixel-wise features across all feature maps in a spatial attention block design.
Moreover, to the best of our knowledge, the CCP has not been adopted as an independent pixel-wise feature extraction descriptor, plugged at the end of DNNs for image classification.

According to the above survey, this paper draws the following conclusions: 1) Existing efforts mainly focused on SP or CCP design, respectively. Unfortunately, no previous work has systematically analyzed the difference between these two pooling operators. 2) These SP and CCP operators only independently extracted spatial or pixel-wise features from feature maps, inevitably losing sight of subtle or salient feature representations in the pooling process. Moreover, these two feature representation types may significantly affect medical image classification results and confidence calibration from the aspect of representation learning, which has yet to be studied before. \textit{Therefore, this paper is the first to rethink the inherent difference between SP and CCP operators through designing a dual-view framework. Based on this new framework, we develop a novel DVPP to aggregate multi-scale dual-view features from the high-level feature maps, involving both multi-scale spatial features and multi-scale pixel-wise features, which is different from previous pooling operators by nature.}

\subsection{DNN Calibration}
Modern DNNs are often overconfident in making wrong predictions \cite{guo2017calibration, zhong2021mislas,patra2023calibrating,zhu2022rethinking}, which is an essential issue in the field of medical computer-aided diagnosis. Recently, a number of works have been proposed to tackle the overconfidence issue of DNNs by calibrating their predicted confidence, and significant progress has been achieved. These calibration methods can generally be classified into regularized training and post-hoc calibration. Regularized training methods aim to improve the confidence calibration of DNNs during training. Thulasidasan et al. ~\cite{NEURIPS2019_36ad8b5f} trained DNNs with the mixup strategy to improve confidence calibration. Mukhot et al.~\cite{mukhoti2020calibrating} applied focal loss to improve network calibration. Other methods, such as Bayesian deep neural networks, deep ensemble networks, and MC-dropout, also have been developed to improve the network calibration through uncertainty estimation \cite{jungo2019assessing, izmailov2020subspace, malinin2019ensemble, wilson2020bayesian}. In contrast, post-hoc calibration methods focus on obtaining well-calibrated results by conducting post-processing on the predicted outputs of DNNs. Temperature scaling \cite{wilson2020bayesian} is the well-accepted post-hoc calibration method by applying a scalar parameter to recalibrate the predicted probabilities. Meta-cal \cite{ma2021meta} integrated the merits of bipartite ranking and selective classification methods for enhancing calibration performance.

Interestingly, we discover that the main objective of these calibration methods is to guide DNNs to emphasize informative features and suppress redundant ones, from the aspect of representation learning, in essence. In other words, they expect the aggregated spatial features from the high-level feature maps to be informative through the SP operator. Subsequently, the classifier of DNNs takes these spatial features as input to generate precise and reliable predictions, but also overlooks pixel-wise feature extraction. \textit{Similarly, our DVPP aggregates multi-scale dual-view features, providing richer and more nuanced feature representations. Therefore, our method has great potential to improve the classification performance and reliability of medical image classification tasks, which have been overlooked in previous efforts.}

\section{Methodology}
\label{sec3}
In this section, we first rethink the difference between SP and CCP operators by designing a novel dual-view framework. Then, this paper introduces the general DVPP form and its five kinds of parameter-free implementation in detail.

\subsection{Rethinking Spatial Pooling and Cross-Channel Pooling with Dual-View Framework}

\begin{figure}
\centering
\includegraphics[width=0.95\linewidth,height=4.5cm]{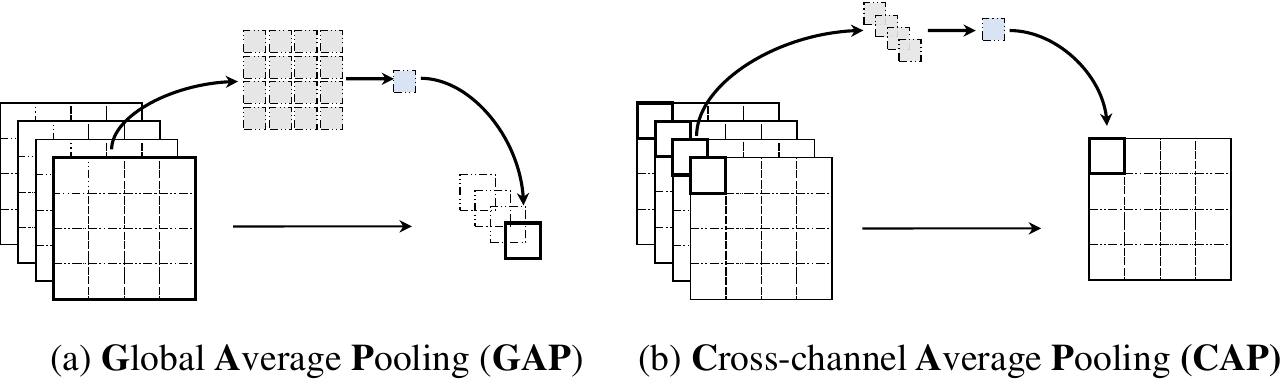}
\caption{The simple implementations of GAP (a) and CAP (b) operators.}
\label{fig2}
\end{figure}

Given the high-level feature maps $X \in R^{C \times H \times W}$ of DNNs, we can apply SP and CCP operators to aggregate spatial features and pixel-wise features, respectively ($C$ denotes the channel; height and width denote the high-level feature map resolution). Here, we adopt GAP and CAP operators as representative examples to illustrate the difference between these two pooling operators. Fig.~\ref{fig2} offers simple implementations of GAP and CAP, which can help readers quickly grasp their concepts within a unique dual-view framework. This paper introduces GAP and CAP step by step in the following section

\textbf{GAP:} As presented in Fig.~\ref{fig2}(a), the global average spatial features extracted from the high-level feature maps through GAP operator are 1D, as introduced in Section~\ref{sec1}. Theoretically, GAP computes the global average spatial feature $\mu_{c}$ of $c-$th feature map across the spatial dimensions, which is formulated as follows:
\begin{equation}
 \mu_{c}= \frac{1}{H \times W}\sum_{i=1}^{H}\sum_{j=1}^{W}x_{c}(i,j). 
 \label{eq:1}
\end{equation}
Global average spatial features of other feature maps also can be obtained with Eq.~\ref{eq:1}.
 
\textbf{CAP:} In contrast to the GAP, pixel-wise average features aggregated by CAP are 2D, as shown in Fig.~\ref{fig2}(b). Given the specific pixel position $(i,j)$, CAP calculates the pixel-wise average feature $\mu(i,j)$ across all channels as follows:
\begin{equation}
 \mu(i,j)= \frac{1}{C}\sum_{k=1}^{C}x(k,i,j). 
\label{eq:2}
\end{equation}
Similarly, pixel-wise average features of other pixel locations can be obtained by Eq.~\ref{eq:2}.

Although a number of improved SP and CCP operators have been developed to aggregate other complex spatial features and pixel-wise features accordingly, no prior effort has combined these two feature types within a dual-view framework for image classification, plugged at the end of DNNs. Seeing that the spatial features and pixel-wise features independently contain salient and subtle feature representations, this paper is the first time to argue that these two feature types not only affect classification results but also impact the model calibration from the aspect of representation learning. \textbf{This distinctive perspective provides a new paradigm to boost classification performance and confidence calibration simultaneously.}


\subsection{Dual-view Pyramid Pooling}
\label{sec:DVP}

\subsubsection{General Form}
\begin{figure*}
\centering
\centerline{\includegraphics[width=1.0\linewidth,height=6.5cm]{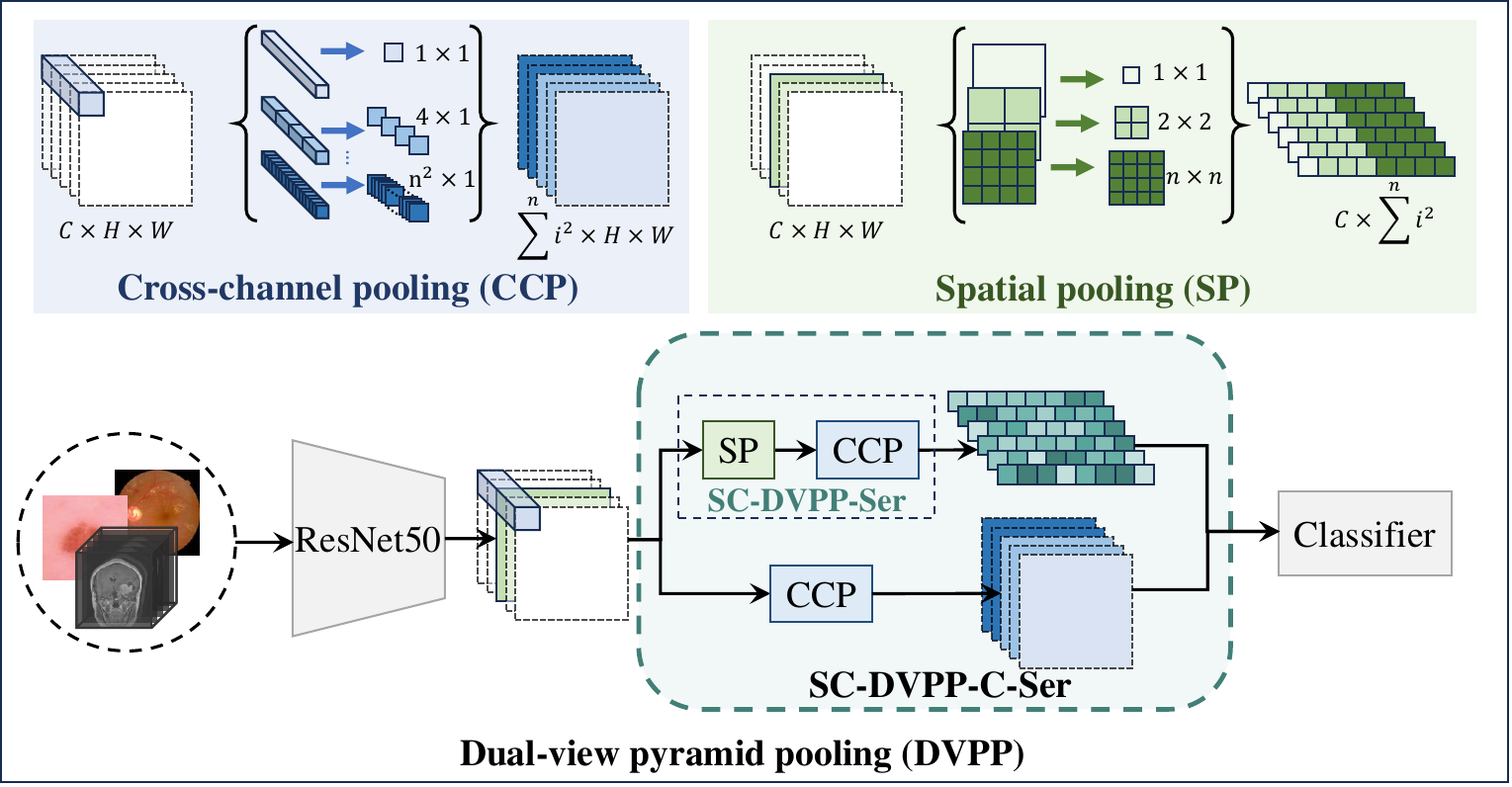}}
  \caption{An representative implementation of our proposed dual-view pyramid pooling (DVPP). Here, we take the SC-DVPP-C-Ser as the example to illustrate DVPP.
  Deep neural networks (DNNs) first take 2D/3D medial images as inputs and generate high-level feature maps. Next, DVPP performs the dual-view pyramid pooling operations to aggregate multi-scale dual-view features from these feature maps, which are fed into the classifier directly. Finally, the classifier generates precise and reliable predicted results. }
	\label{fig3}
\end{figure*}

To meet the expectations of both classification and confidence calibration performance boosting based on the proposed dual-view framework. We propose a dual-view pyramid pooling (DVPP) operator to aggregate multi-scale dual-view features from the high-level feature maps of DNNs. Note that several multi-scale dual-view feature types can be obtained via complex DVPP design, but those are not the focus of this paper. Hence, we only utilize the DVPP operator to aggregate multi-scale average dual-view features from a dual-axis perspective, plugging at the end of DNNs for medical image classification and confidence calibration directly, as illustrated in Fig.~\ref{fig3}. This paper formulates the general form of our proposed DVPP as follows:
\begin{equation}
Z_{DVPP}= f_{DVPP}(X), 
\label{eq:3}
\end{equation}
where $Z_{DVPP}$ denotes the multi-scale dual-view features, and $f_{DVPP}(X)$ denotes the dual-view pyramid pooling operations implemented on high-level feature maps $X$, which inherits advantages of SP and CCP operators. Here, SP and CCP indicate spatial pyramid pooling and cross-channel pyramid pooling operations. Considering that parametric DVPP implementations do not work well based on the experimental results, we focus on introducing five kinds of parameter-free DVPP implementation.

\subsubsection{Parameter-free DVPP Implementation}
\label{sec:3.21}

\begin{figure*}
\centering
\centerline{\includegraphics[width=0.95\linewidth, height=3.5cm]{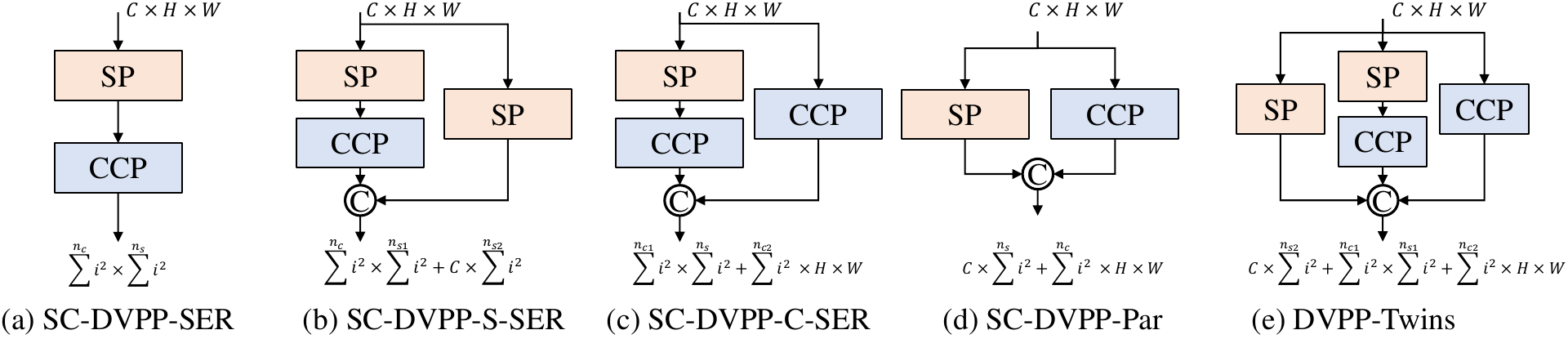}}
  \caption{Five representative parameter-free DVPP implementations: SC-DVPP-Ser, SC-DVPP-S-Ser, SC-DVPP-C-Ser, SC-DVPP-Par, and Twins-DVPP.}
	\label{fig4}
\end{figure*}

Under the paradigm of parameter-free DVPP implementation, there is a significant problem of how to effectively aggregate multi-scale dual-view features via our proposed DVPP, involving multi-scale spatial features and multi-scale pixel-wise features . Therefore, to fulfill our proposed DVPP, this paper designs five parameter-free DVPP implementations: SC-DVPP-Ser, SC-DVPP-S-Ser, SC-DVPP-C-Ser, SC-DVPP-Par, and Twins-DVPP, as provided in Fig.~\ref{fig4}. \textbf{To be specific, SC-DVPP-Ser performs the SP and CCP operators serially, while SC-DVPP-Par performs the SP and CCP operators in parallel.}

\textbf{SC-DVPP-Ser:} We first utilize the SP operator to aggregate multi-scale spatial features from high-level feature maps. Then, CCP is applied to extract multi-scale dual-view features from aggregated multi-scale spatial features. This paper formulates the \emph{SC-DVPP-Ser} as follows:
\begin{equation}
Z_{\textit{SC-Ser}} = f_{CCP}(f_{SP}(X)), 
\label{eq:3}
\end{equation}
$Z_{\textit{SC-Ser}}$ denotes the extracted multi-scale dual-view features (Differing from concepts of multi-scale spatial features and multi-scale pixel-wise features), and $f_{SP}$ denotes SP operations across spatial dimensions, and $f_{CCP}$ denotes CCP operations across all channels. Fig.~\ref{fig4}(a) presents a simple implementation of SC-DVPP.
\textit{In particular, there is another implementation of SC-DVPP-Ser: $Z_{\textit{SC-Ser}}= f_{SP}(f_{CP}(X))$, which first performs the CCP operator and is followed by the SP operator. These two implementations of SC-DVPP-Ser are equivalent theoretically.}


\textbf{SC-DVPP-S-Ser:} Considering that the final aggregated features aggregated by \em{SC-DVPP-Ser} operator are multi-scale dual-view features, lacking original multi-scale spatial features. To address this problem, we add another SP branch for the \em{SC-DVPP-Ser} to aggregate multi-scale spatial features from high-level feature maps independently, termed \em{SC-DVPP-S-Ser}. The hybrid multi-scale dual-view features $Z_{SC-S-Ser}$ via \em{SC-DVPP-S-Ser} operator are expressed as:
\begin{equation}
Z_{\textit{SC-S-Ser}} = [f_{CCP}(f_{SP}(X)),f_{SP}(X)].
\label{eq:4}
\end{equation}

Fig.~\ref{fig4}(b) provides an implementation example of SC-DVPP-S-Ser, a combination of SC-DVPP and SP operations.

\textbf{SC-DVPP-C-Ser:} In contrast to SC-DVPP-S-Ser, SC-DVPP-C-Ser adds another CCP branch for SC-DVPP-Ser to extract pure multi-scale pixel-wise features, as shown in Fig.~\ref{fig4}(c). The hybrid multi-scale dual-view features $Z_{SC-C-Ser}$ via CS-DVPP-C operator are written as follows:
\begin{equation}
Z_{SC-C-Ser} = [f_{CCP}(f_{SP}(X)), f_{CCP}(X)].
\label{eq:5}
\end{equation}

\textbf{SC-DVPP-Par:} Different from SC-DVPP-Ser, SC-DVPP-Par aggregates parallel multi-scale dual-view features $Z_{SC-Par}$  from high-level feature maps concurrently based on SP and CCP operators, which are the combination of multi-scale spatial features and multi-scale pixel-wise features, as listed in Fig.~\ref{fig4}(d). We formulate the SC-DVPP-Par operator as follows:
\begin{equation}
Z_{\textit{SC-Par}} = [f_{SP}(X),f_{CCP}(X)].
\label{eq:6}
\end{equation}

\textbf{Twins-DVPP:} Unlike the above four kinds of DVPP implementation, Twins-DVPP attempts to integrate the merits of SC-DVPP-Ser and SC-DVPP-Par, as presented in Fig.~\ref{fig4}(f). Therefore, we formulate the hybrid multi-scale dual-view features $Z_{Twins}$ via Twins-DVPP operator as follow:
\begin{equation}
Z_{Twins}= [f_{SP}(f_{CCP}(X)), f_{CCP} (X), (f_{SP}(X)].
\label{eq:7}
\end{equation}

\textbf{Analysis of Parameter-Free DVPP:} The fundamental motivation to design DVPP operator for obtain multi-scale dual-view features from high-level feature maps is that these two features contain both salient and subtle representation information, which existing pooling methods cannot obtain. These feature types clearly affect classification and confidence calibration performance according to our analysis.

Moreover, the multi-level pooling selections of the SP and CCP operators significantly impact the performance of the five parameter-free DVPP implementations. This paper focuses on the multi-level pooling selections of SP and CCP for SC-DVPP-Ser based on ablation experiments. SC-DVPP-S-Ser, SC-DVPP-C-Ser, and SC-DVPP-Par are implemented by selecting the top two best multi-level pooling settings of SP and CCP in SC-DVPP-Ser. Twins-DVPP is implemented by using both SC-DVPP-Ser and SC-DVPP-Par.

\section{Experiments}
\label{sec:exper}

\subsection{Datasets} 
In this paper, we utilize six 2D/3D publicly available medical image datasets to examine the effectiveness of our proposed method.

\textbf{ISIC2018 \cite{tschandl2018ham10000}.} It is a skin lesion dataset with 10,015 images of seven classes. In order to make a fair comparison, we follow the same dataset splitting and data prepossessing strategies in literature~\cite{Zhang2023ppcr}.

\textbf{BTM~\cite{BTM}.}  It is a publicly available magnetic resonance imaging (MRI) dataset of the brain tumor. It contains 3,254 MRI images with four tumor types: benign, meningioma, glioma, and pituitary. The BTM consists of training (2,870 images) and testing (384 images) subsets. In particular, 20 \% of training images are adopted as the validation subset for selecting the best-trained model.

\textbf{APTOS2019~\cite{APTOS}.} It is a public retinal fundus image dataset of diabetic retinopathy (DR). It contains 3,662 images with five different severity levels of DR. This paper splits it into three disjoint subsets: training (2,510), validation (608), and testing (544).

\textbf{NIH-CXR-LT~\cite{holste2022long}}. It is a long-tailed Chest X-ray image dataset of thorax diseases. The dataset contains 88,637 images with 19 thorax diseases and a normal class. Here, we follow the same dataset splitting strategy used in \cite{holste2022long} for a fair comparison.

\textbf{OASIS~\cite{wen2020convolutional}}. It is an MRI dataset of Alzheimer’s disease (AD), including 193 participants aged 62 years or older. Following the same preprocessing pipeline in~\cite{wen2020convolutional}, 76 cognitively normal subjects (CN) and 78 AD subjects were selected for experiments. We split them into three disjoint subsets: training (104), validation (26), and testing (24) based on the participant level.


\textbf{ABIDE-I~\cite{di2014autism}}. It is a resting-state functional magnetic resonance imaging (R-fMRI) dataset of autism spectrum disorder (ASD), comprised of 539 ASD patients and 573 normal subjects. After the preprocessing, 871 subjects were selected for the following experiments. This paper also splits them into three disjoint subsets: training (592), validation (131), and testing (148) based on the subject level.


\subsection{Experimental Settings}
This paper implements our DVPP, SOTA pooling, and calibration methods based on Pytorch and Python. We run all experiments on a NVIDIA RTX A6000 GPU. For four 2D medical image datasets, the initial learning rate, training epochs, and batch size are set to 0.002, 150, and 32, respectively. We resize the input size of images into $224 \times 224$.
The stochastic gradient descent (SGD) optimizer is utilized to update the parameters of DNNs during training. Specifically, this paper decreases the learning rate by a factor of 10 every ten epochs; when the epochs are over 100, it is decreased by a factor of 5 every 20 epochs. During training, we utilize standard data augmentation methods to augment training data. We use the same experiment settings in literature~ \cite{wen2020convolutional} for the OASIS and ABIDE-I. The code of this paper is available at \href{https://github.com/Tloops/DVPP}{https://github.com/Tloops/DVPP}.

\subsection{Evaluation Metrics}
To evaluate the classification results of our DVPP, comparable pooling methods and calibration methods, especially considering the data imbalance issue, we adopt four commonly used metrics: accuracy (ACC), balanced accuracy (bAcc), macro F1 (mF1), and kappa value. Additionally, to measure calibration performance of DNNs, we adopt the expected calibration error (ECE) and Brier score (BS)~\cite{nixon2019measuring} as evaluation measures.

\subsection{Baselines} 
In this paper, we not only adopt FC layers as the comparable method but also utilize the following pooling methods to test the effectiveness of our proposed DVPP, which can be classified as:
\begin{itemize}
    \item Parametric pooling: LIP, S3Pool, AdaPool, UPool, and MP. 
    \item Parameter-free pooling: GAP, GMP, SPP, CAP, and CCPP.
\end{itemize}

Moreover, we also adopt competitive calibration methods for comparison: temperature scaling (Temp.), focal loss (FL)~\cite{lin2017focal}, LADM \cite{cao2019learning}, and Mixup~\cite{zhang2018mixup}.

\subsection{Ablation Study}
In this section, we conduct a series of ablation experiments to choose the proper multi-level pooling settings for our five kinds of parameter-free DVPP implementation based on ISIC2018 and BTM. Here, we take classical ResNet18 and ResNet50 as backbone networks.

\subsubsection{Effects of Multi-level Pooling Selection in SC-DVPP-Ser}

Table~\ref{tab:1} offers the medical image classification results of different multi-level pooling settings of SP and CCP in SC-DVPP-Ser. Specifically:
\begin{itemize}
    \item When the multi-level pooling of SP and CCP is set to 1 and 0, it is equivalent to GAP.
   \item When the multi-level pooling of SP and CCP is set to 0 and 1, it is equivalent to CAP.
    \item When the multi-level pooling of SP and CCP is set to $\{2, 3, 4\}$ and 0, it denotes SPP.
    \item When the multi-level pooling of SP and CCP is set to 0 and $\{2, 3, 4\}$, it denotes CCPP
\end{itemize}

The results show marked differences in medical image classification performance among GAP, SPP, CAP, and CCPP, confirming that the features aggregated by each pooling operator have unique impacts on classification results, consistent with our hypothesis. SPP based on the multi-level pooling setting $2$ of SP, achieves better performance than other multi-level pooling settings of SP. Similarly, CCPP under the multi-level pooling setting $3$ of CCP,  generally performs better than other multi-level pooling settings of CCP. Hence, these two multi-level pooling settings of SP and CCP for SSP and CCPP are adopted for subsequent comparisons.

According to Table~\ref{tab:1}, we see visible performance fluctuations among SC-DVPP-Ser, implying that choosing the best multi-level pooling setting for SC-DVPP-Ser is challenging. Moreover, SC-DVPP-Ser with the multi-level pooling settings $\{3, 4\}$ and $\{4, 2\}$ of SP and CCP generally performs better than that with other multi-level pooling settings. Therefore, we adopt these two multi-level pooling settings of SC-DVPP-Ser for the ablation experiments of SC-DVPP-S-Ser and SC-DVPP-C-Ser. The multi-level pooling settings experiments of SC-DVPP-Par are conducted by selecting six multi-level pooling settings of SC-DVPP-Ser.

\begin{table*}
\caption{Performance comparisons of multi-level pooling settings of SP and CPP in SC-DVPP-Ser.}
\label{tab:1}
\center
\begin{tabular}{cc|cc|cc|cc|cc}
\hline
  \multicolumn{2}{c|}{\multirow{2}{*} {Multi-level Pooling}} &\multicolumn{4}{c}{ISIC2018} &\multicolumn{4}{c}{BTM} \\
  \multicolumn{2}{c|}{}  &\multicolumn{2}{c|}{ReNet18} &\multicolumn{2}{c}{ResNet50} &\multicolumn{2}{c|}{ReNet18} &\multicolumn{2}{c}{ResNet50} \\ 
 \hline
    SP&CCP&ACC  &mF1   &ACC  &mF1   &ACC  &mF1   &ACC  &mF1\\ 
    0&1   &83.85&54.97&83.85&56.92&84.12&83.45&84.64&83.74\\
    0&2   &80.73&52.58&84.90&77.51&83.85&83.08&83.33&82.58\\
    0&3   &80.21&51.65&80.21&73.05&83.33&82.52&85.42&84.72\\
    0&4   &83.85&68.68&85.94&66.18&83.00&82.28&81.98&80.51\\
    1&0   &80.21&64.31&81.77&59.70&85.16&84.54&83.33&82.28\\
    1&1   &78.65&37.11&80.21&39.12&84.12&83.79&83.59&83.30\\
    1&2   &80.21&49.24&81.77&51.53&84.12&83.40&82.03&81.05\\
    1&3   &83.33&52.23&86.46&82.02&85.94&85.57&83.85&83.16\\
    1&4   &82.81&54.88&82.29&63.37&82.29&81.06&85.42&84.99\\
    2&0   &81.77&69.67&83.33&63.09&83.07&82.38&84.64&83.96\\
    2&1   &82.81&45.60&83.85&50.28&85.16&84.77&85.94&85.48\\ 
    2&2   &84.38&59.04&82.29&57.22&84.38&83.67&86.20&85.72\\
    2&3   &83.33&53.17&85.94&\textbf{82.03}&84.90&84.34&86.20&85.95\\
    2&4   &82.29&51.82&80.73&73.81&85.42&84.90&83.85&82.93\\
    3&0   &83.85&61.36&81.77&71.82&82.81&82.03&85.94&85.40\\
    3&1   &83.33&55.03&80.21&53.59&85.68&85.26&85.42&84.93\\
    3&2   &80.73&50.61&82.81&59.19&81.25&79.81&84.12&83.01\\
    3&3   &82.29&68.08&82.81&59.76&81.77&80.80&85.68&84.88\\
    3&4   &84.90&\textbf{73.91}&\textbf{86.98}&81.33&86.46&86.20&86.72&86.13\\
    4&0   &81.25&54.33&81.77&70.97&75.26&74.21&83.85&83.24\\
    4&1   &83.85&56.07&82.81&72.78&85.68&85.34&84.64&83.78\\
    4&2   &\textbf{85.42}&63.61&86.98&70.08&\textbf{86.72}&\textbf{86.31}&\textbf{86.98}&\textbf{86.57}\\
    4&3   &83.33&70.92&81.77&62.11&83.33&82.54&86.20&85.77\\
    4&4   &80.21&63.90&84.38&75.47&81.22&79.97&81.22&80.13\\
   \hline
\end{tabular}
\end{table*}

\subsubsection{Effects of Multi-level Pooling Selection in SC-DVPP-S-Ser}

Table~\ref{tab:2} presents the classification results of multi-level pooling settings $\{SP, CCP, SP\}$ in SC-DVPP-S-Ser. As described previously, the implementations of SC-DVPP-S-Ser based on the top two best multi-level pooling settings of SC-DVPP-Ser. Interestingly, we note that SC-DVPP-S-Ser generally performs worse than the original SC-DVPP-Ser in our experiments. The possible reason behind this phenomenon is that SC-DVPP-Ser effectively extracts multi-scale dual-view features, involving abundant spatial features. Unfortunately, the additional SP operator also extracts multi-scale spatial features, introducing redundant spatial feature information, thereby leading to performance degradation.

\begin{table*}
\caption{Performance comparisons of multi-level pooling settings of SP and CCP in SC-DVPP-S-Ser.}
\label{tab:2}
\center
\begin{tabular}{cc|c|cc|cc|cc|cc}
		\hline
  \multicolumn{3}{c|}{\multirow{2}{*} {Multi-level Pooling}} &\multicolumn{4}{c|}{ISIC2018} &\multicolumn{4}{c}{BTM} \\
  \multicolumn{3}{c|}{}  &\multicolumn{2}{c|}{ReNet18} &\multicolumn{2}{c|}{ResNet50} &\multicolumn{2}{c|}{ReNet18} &\multicolumn{2}{c}{ResNet50} \\ 
 \hline
    SP&CCP&SP&ACC  &mF1   &ACC  &mF1   &ACC  &mF1   &ACC  &mF1\\  
    3&4&0&84.90&73.91&\textbf{86.98}&\textbf{81.33}&86.46&86.20&86.72&86.13\\
    3&4&1&83.33&60.65&84.90&77.92&83.33&82.51&84.90&84.39\\
    3&4&2&82.29&\textbf{75.07}&83.85&79.89&83.33&82.40&85.42&85.02\\
    3&4&3&81.25&46.94&80.21&60.12&83.07&82.20&83.33&82.64\\
    3&4&4&80.21&58.81&84.90&72.88&80.47&79.39&83.33&82.41\\
    4&2&0&85.42&63.61&86.98&70.08&\textbf{86.72}&\textbf{86.31}&\textbf{86.98}&\textbf{86.57}\\
    4&2&1&\textbf{85.42}&69.25&84.38&74.12&83.07&82.20&83.85&83.28\\
    4&2&2&82.29&54.45&84.38&78.93&82.55&81.61&84.12&83.26\\
    4&2&3&82.29&70.33&81.25&59.05&82.03&80.87&84.64&84.08\\
    4&2&4&80.73&60.31&84.90&78.16&81.77&80.19&83.07&82.36\\
  \hline
\end{tabular}
\end{table*}

\subsubsection{Effects of Multi-level Pooling Selection in SC-DVPP-C-Ser}

Table~\ref{tab:3} lists the result comparisons of multi-level pooling settings for SC-DVPP-C-Ser. Similarly, the implementations of SC-DVPP-C-Ser also under the two best multi-level pooling settings for SC-DVPP-Ser. SC-DVPP-C-Ser with the multi-level pooling setting $\{4, 2, 3\}$ of SP, CCP and CCP obtains better performance than it with other multi-level pooling settings. Unlike the performance of SC-DVPP-S-Ser, which is slightly worse than the original SC-DVPP-Ser, SC-DVPP-C-Ser achieves better performance than SC-DVPP-Ser in general. A convincing reason to explain the phenomenon is that extracted multi-scale dual-view features by SC-DVPP-Ser do not contain enough multi-scale pixel-wise features, and the independent CCP operator is complementary. 

\begin{table*}
\caption{Performance comparisons of multi-level pooling settings of SP, CCP, and CCP in SC-DVPP-C-Ser.}
	\label{tab:3}
	\center
	\begin{tabular}{cc|c|cc|cc|cc|cc}
		\hline
  \multicolumn{3}{c|}{\multirow{2}{*} {Multi-level Pooling}} &\multicolumn{4}{c|}{ISIC2018} &\multicolumn{4}{c}{BTM} \\
  \multicolumn{3}{c|}{}  &\multicolumn{2}{c|}{ReNet18} &\multicolumn{2}{c|}{ResNet50} &\multicolumn{2}{c|}{ReNet18} &\multicolumn{2}{c}{ResNet50} \\ 
 \hline
    SP&CCP&CCP&ACC  &mF1   &ACC  &mF1   &ACC  &mF1   &ACC  &mF1\\  
    3&4&0&84.90&\textbf{73.91}&86.98&81.33&86.46&86.20&86.72&86.13\\
    3&4&1&81.77&55.68&87.50&81.34&84.64&83.79&85.16&84.80\\
    3&4&2&83.85&58.61&84.90&76.34&83.85&83.24&82.55&81.66\\
    3&4&3&83.33&56.42&85.94&66.35&83.59&82.83&85.16&84.59\\
    3&4&4&80.21&55.10&83.85&73.45&82.03&80.99&85.68&85.19\\
    4&2&0&85.42&63.61&86.98&70.08&86.72&86.31&86.98&86.57\\
    4&2&1&85.42&55.73&83.85&62.94&83.33&82.46&85.68&85.15\\
    4&2&2&83.85&51.09&83.33&60.88&84.12&83.41&85.68&85.08\\   4&2&3&\textbf{85.94}&73.15&\textbf{88.02}&\textbf{82.91}&\textbf{87.24}&\textbf{86.82}&\textbf{87.76}&\textbf{87.44}\\
    4&2&4&81.77&62.93&83.33&60.68&85.68&85.07&86.46&85.94\\
  \hline
\end{tabular}
\end{table*}

\subsubsection{Effects of Multi-level Pooling Selection in SC-DVPP-Par}
Table~\ref{tab:4} offers the classification results of SC-DVPP-Par with different multi-level pooling settings. It can be observed that SC-DVPP-Par with the multi-level pooling setting $\{1,3\} $ generally performs better than that with other settings. Compared to the performance of SC-DVPP-Ser, SC-DVPP-Par slightly performs worse. We explain inherent reasons behind the results as follows: 1) SC-DVPP-Par extracts pure multi-scale spatial features and multi-scale pixel-wise features, which may contain many redundant feature representations, harming the performance of DNNs. 2) SC-DVPP-Ser aggregates multi-scale dual-view features that involve enough feature representations to allow DNNs to get promising performance.

\begin{table*}
\caption{Performance comparisons of multi-level pooling settings of SP and CCP in SC-DVPP-Par.}
	\label{tab:4}
	\center
	\begin{tabular}{c|c|ccccccccc}
		\hline
  \multicolumn{2}{c|}{\multirow{2}{*} {Multi-level Pooling}} &\multicolumn{4}{c|}{ISIC2018} &\multicolumn{4}{c}{BTM} \\
  \multicolumn{2}{c|}{}  &\multicolumn{2}{c|}{ReNet18} &\multicolumn{2}{c|}{ResNet50} &\multicolumn{2}{c|}{ReNet18} &\multicolumn{2}{c}{ResNet50} \\ 
 \hline
    SP&CCP&ACC  &mF1   &ACC  &mF1   &ACC  &mF1   &ACC  &mF1\\  
    1&3&\textbf{82.29}&\textbf{72.38}&85.42&79.05&82.55&81.44&85.68&85.18\\
    2&3&82.29&55.52&\textbf{86.98}&\textbf{81.72}&\textbf{84.12}&\textbf{83.36}&83.33&82.34\\
    3&1&82.29&67.83&84.90&78.32&83.59&82.86&82.03&80.67\\
    3&4&80.21&53.95&83.33&69.28&82.55&81.43&\textbf{86.20}&\textbf{85.78}\\
    4&2&82.29&49.07&84.37&73.60&81.51&80.66&83.59&82.88\\
    4&3&81.25&71.86&83.85&76.91&80.73&79.63&80.73&80.64\\
  \hline
\end{tabular}
\end{table*}

\subsubsection{Effects of Multi-level Pooling Selection in Twins-DVPP}

The implementations of Twins-DVPP by selecting the best two multi-level pooling settings of SC-DVPP-Ser and SC-DVPP-Par. Table~\ref{tab:5} presents four Twins-DVPP implementations. Sad to say, the classification performance of Twins-DVPP is worse than that of other four kinds of DVPP implementation. The key reason to explain the results in Table~\ref{tab:5} is that the extracted dual-view multi-scale feature via the Twins-DVPP operator contain redundant feature representations, inevitably weakening the performance.

\begin{table*}
\caption{Performance comparisons of multi-level pooling settings of SP and CCP in Twins-DVPP}
	\label{tab:5}
	\center
		\begin{tabular}{cc|c|c|ccccccccc}
		\hline
  \multicolumn{4}{c|}{Multi-level} &\multicolumn{4}{c|}{ISIC2018} &\multicolumn{4}{c}{BTM} \\
  \multicolumn{2}{c|}{SC-DVPP-Ser} &\multicolumn{2}{c|}{SC-DVPP-Par} &\multicolumn{2}{c|}{ReNet18} &\multicolumn{2}{c|}{ResNet50} &\multicolumn{2}{c|}{ReNet18} &\multicolumn{2}{c}{ResNet50} \\ 
  \hline
    SP&CCP&CCP&SP&ACC  &mF1   &ACC  &mF1   &ACC  &mF1   &ACC  &mF1\\  
    3&4&2&3&82.29&51.39&84.90&64.08&82.81&81.78&84.38&83.85\\
    3&4&3&4&\textbf{82.81}&\textbf{71.32}&80.73&61.38&\textbf{84.90}&\textbf{84.26}&\textbf{85.16}&\textbf{84.67}\\
    4&2&2&3&81.77&69.71&\textbf{85.42}&\textbf{72.33}&83.33&82.28&85.16&84.64\\
    4&2&3&4&81.25&47.95&82.81&73.20&82.81&81.87&83.59&82.81\\
  \hline
\end{tabular}
\end{table*}

According to Table~\ref{tab:1}-Table~\ref{tab:5}, we conclude as follows: 1) The choice of multi-level pooling settings of all five kinds of DVPP implementation have significant impacts on medical image classification results. We can see that it is challenging to choose the proper multi-level pooling settings for them. 2) Adding another pure SP branch in SC-DVPP-S-Ser worsens the general performance of SC-DVPP-Ser. In contrast, adding another pure CCP branch in SC-DVPP-C-Ser slightly enhances the performance of SC-DVPP-Ser. The possible reason for explaining this phenomenon is that the number of spatial features aggregated by SP operations is more than the number of pixel-wise features aggregated by CCP operations, unavoidably introducing redundant feature representations that harm medical image classification performance.3) Aggregating more multi-scale dual-view features does not further classification performance due to redundant feature representations. 4) Based on the results of five kinds of DVPP implementation, SC-DVPP-Ser with multi-level pooling $\{3,4 \}$ and SC-DVPP-C-Ser with multi-level pooling $\{4,2,3 \}$ are adopted as the representative DVPP implementations for the following comparisons.

Furthermore, our DVPPs under ResNet50 perform better than those based on ResNet18. Hence, ResNet50 is adopted as a backbone for verifying our DVPP's effectiveness through comparisons to competitive pooling methods.

\subsection{Comparisons with SOTA Pooling Methods}

\subsubsection{Validation on 2D Medical Image Datasets}

\textbf{Results on ISIC2018.} Table~\ref{tab:6}(Left) offers the medical classification results of two representative DVPP implementations, competitive pooling methods, and FC on the ISIC2018 dataset. Here, we take pre-trained ResNet50, VGG16, and Swin Transformer (Swin-T) as backbone networks to validate the effectiveness and generalization ability of our method. Across all three backbone networks, our two DVPP implementations (SC-DVPP-Ser and SC-DVPP-C-Ser) generally achieves better medical classification and confidence calibration performance than other advanced pooling methods and FC, and the performance of SC-DVPP-Ser is second only to SC-DVPP-C-Ser.
For example, SC-DVPP-C-Ser surpasses FC and parametric pooling methods based on ResNet50, e.g., LIP, UPool, and MP, by above absolute 1.74\% of bAcc and 2.18\% of mF1; while obtaining 0.44\% decrease of ECE. Noticeably, compared with parameter-free pooling methods, particularly SPP and CCPP, our SC-DVPP-C-Ser significantly outperforms them by over absolute \textbf{6.51\%} of bAcc and \textbf{4.19\%} of mF1, while also reducing ECE by 0.77\% and BS by \textbf{2.4\%} correspondingly. The results demonstrate the superiority of our DVPP in exploiting multi-scale dual-view features, which boosts medical classification and confidence calibration performance.

\textbf{Results on BTM.} Table~\ref{tab:6}(Right) presents the medical image classification and confidence calibration results of DVPP, competitive pooling methods, and FC on the BTM dataset. Similar to the ISIC2018 dataset, we also adopt three backbone networks to verify the effectiveness and generalization ability of DVPP. Our two DVPP implementations also generally perform better than other parametric pooling and parameter-free pooling methods in terms of brain tumor classification and confidence calibration. Notably, under ResNet50, SC-DVPP-C-Ser surpasses LIP, AdaPool, and UPool by above absolute \textbf{4.85\%} in accuracy, bAcc, mF1, and kappa, while reducing ECE and BS by more than 1.7\% and \textbf{6.08\%}. Compared to SPP under VGG16, SC-DVPP-C-Ser gains more than \textbf{3.85\%} in the four classification measures and reduces two confidence calibration measures by \textbf{4.14\%}. The experimental results on the BTM also demonstrate that the multi-scale dual-view features aggregated by DVPP significantly boost both classification and confidence calibration performance.

Overall, the results in Table~\ref{tab:6} demonstrate the effectiveness and generalization ability of our DVPP over competitive pooling methods and FC in terms of classification and confidence calibration across three representative backbone networks, consistent with our expectations. In other words, the aggregated multi-scale dual-view features obtained by DVPP, involving both salient and subtle features, significantly affecting medical image classification and confidence calibration performance. Additionally, we use the pre-trained ResNet50 as the backbone network in subsequent experiments to further prove the effectiveness and generalization ability of the proposed DVPP on the other two imbalanced medical image datasets.

\begin{table*}
    \centering
    \caption{Performance comparisons of different pooling methods and FC in terms of medical image classification and confidence calibration on ISIC2018 and BTM. Note that \protect\usym{2714} and \protect\usym{2718} denote parameter-free pooling paradigm and parameteric pooling paradigm accordingly.}
    \label{tab:6}
  \resizebox{0.99\columnwidth}{!}{
    \begin{tabular}{c|c|c|cccccc|cccccc} 
        \toprule
        \multirow{2}{*}{Backbone} &\multirow{2}{*}{Method}&\multirow{2}{*}{Paradigm}&\multicolumn{6}{c}{ISIC2018}&\multicolumn{6}{c}{BTM}\\ 
        && &ACC$\uparrow$ & bAcc$\uparrow$ & mF1$\uparrow$&kappa$\uparrow$& ECE$\downarrow$ & BS$\downarrow$ &ACC$\uparrow$ & bAcc$\uparrow$ & mF1$\uparrow$&kappa$\uparrow$& ECE$\downarrow$ & BS$\downarrow$  \\
        \midrule
       \multirow{13}{*}{ResNet50} 
        &FC&\usym{2718}&78.65&41.40&42.80&58.19&10.57&32.91&84.38&82.95&83.61&78.82&14.67&30.21\\
        &GAP&\usym{2714}&81.77&54.41&59.70&64.15&13.66&31.21&83.33&82.47&82.28&77.43&13.90&29.39\\
        &GMP&\usym{2714}&85.42&74.62&73.12&72.73&10.03&22.80&86.46&85.60&86.04&81.68&12.62&25.33\\
        &SPP&\usym{2714}&85.94&75.55&78.49&74.08&9.47&23.92&84.38&83.06&83.78&78.81&13.92&28.03\\
        &LIP&\usym{2718}&86.46&74.30&73.63&74.51&9.48&\textbf{21.39}&82.81&81.72&81.94&76.68&15.15&30.83\\
        &S3Pool&\usym{2718}&81.77&64.42&59.65&66.89&\textbf{8.22}&28.01&81.25&80.01&80.22&74.54&15.83&32.94\\
    &AdaPool&\usym{2718}&84.38&59.02&61.92&71.55&9.96&26.09&85.68&85.13&85.16&80.65&12.74&26.85\\
        &UPool&\usym{2718}&85.94&80.32&80.62&74.74&9.80&22.17&83.59&82.71&82.70&77.77&13.28&30.55\\
        &MP&\usym{2718}&86.98&77.86&80.57&75.24&9.14&23.52&83.85&82.90&83.07&78.12&13.66&29.62\\ 
        &CAP&\usym{2714}&83.85&54.73&56.92&70.62&10.51&25.50&84.64&83.81&83.74&79.20&12.28&27.13\\ 
        &CCPP&\usym{2714}&84.90&73.41&77.51&71.54&10.07&24.66&83.33&82.63&82.58&77.43&13.58&29.42\\ 
        &SC-DVPP-Ser&\usym{2714}&86.98&78.59&81.33&75.24&9.83&21.63&86.72&86.13&86.38&82.05&\textbf{10.27}&24.01\\    
        &SC-DVPP-C-Ser&\usym{2714}&\textbf{88.02}&\textbf{83.43}&\textbf{82.91}&\textbf{78.17}&8.70&21.52&\textbf{87.76}&\textbf{87.14}&\textbf{87.44}&\textbf{83.47}&11.12&\textbf{22.69}\\  
            \hline  \hline    
        \multirow{13}{*}{VGG16} 
        &FC&\usym{2718}&84.90&55.74&59.82&70.82&9.37&23.56&85.68&85.09&85.20&80.64&13.03&26.52\\
        &GAP&\usym{2714}&85.42&55.77&60.71&71.59&5.37&21.58&85.94&85.44&85.57&81.02&11.63&27.30\\
        &GMP&\usym{2714}&85.94&57.61&61.50&73.42&8.26&22.69&84.12&83.49&83.43&78.48&13.91&29.54\\
        &SPP&\usym{2714}&85.94&60.64&64.39&73.74&7.27&22.12&83.85&82.70&83.18&78.12&13.75&27.83\\
      
        &LIP&\usym{2718}&85.94&59.28&64.31&73.03&9.26&24.13&85.68&85.15&85.32&80.64&13.00&26.69\\
        
        &S3Pool&\usym{2718}&84.38&57.35&60.88&70.73&8.53&24.73&83.33&82.60&82.76&77.41&13.54&29.70\\
        &AdaPool&\usym{2718}&84.90&59.27&63.33&71.62&10.37&25.39&84.64&83.63&83.95&79.20&12.38&26.67\\
        &UPool&\usym{2718}&85.94&59.17&62.56&73.76&8.07&19.60&81.77&80.62&80.68&75.31&14.63&31.14\\
        &MP&\usym{2718}&84.90&57.63&61.33&71.75&6.55&22.49&85.16&84.38&84.47&79.94&11.15&26.14\\ 
        &CAP&\usym{2714}&84.90&56.76&59.36&72.20&6.92&21.96&83.07&82.23&82.08&77.08&10.66&29.37\\
        &CCPP&\usym{2714}&85.42&60.95&64.29&72.24&8.58&24.39&84.38&82.89&83.32&78.84&12.54&28.43\\ 
        &SC-DVPP-Ser&\usym{2714}&86.98&64.35&64.10&76.08&\textbf{4.98}&\textbf{19.02}&86.20&85.19&85.78&81.34&12.92&27.00\\
        &SC-DVPP-C-Ser&\usym{2714}&\textbf{88.02}&\textbf{62.11}&\textbf{66.52}&\textbf{77.26}&5.23&19.17&\textbf{87.50}&\textbf{86.82}&\textbf{87.04}&\textbf{83.10}&\textbf{9.61}&\textbf{21.65}\\
            \hline \hline
        \multirow{13}{*}{Swin-T}
        &FC&\usym{2718}&88.54&83.90&78.48&79.62&6.38&19.40&83.85&82.39&82.93&78.10&13.12&27.57\\ 
        &GAP&\usym{2714}&89.58&84.81&85.67&81.10&6.90&16.87&83.59&82.21&82.77&77.74&14.07&29.57\\
        &GMP&\usym{2714}&89.58&83.66&77.00&81.57&6.36&19.04&82.03&80.24&80.95&75.57&13.63&30.79\\
        &SPP&\usym{2714}&89.06&85.99&86.00&80.46&6.85&17.70&83.33&81.96&82.45&77.39&15.03&31.01\\
        &LIP&\usym{2718}&88.02&82.83&78.22&78.53&8.47&20.78&80.21&78.64&78.75&73.16&14.71&33.08\\
        &S3Pool&\usym{2718}&89.06&85.82&85.65&80.73&\textbf{5.60}&18.62&83.59&82.21&82.75&77.74&14.79&30.56\\
        &AdaPool&\usym{2718}&89.58&83.23&84.34&81.08&6.92&17.78&81.77&80.40&80.41&75.26&15.96&32.51\\
        &UPool&\usym{2718}&89.06&86.08&81.29&80.40&5.23&18.23&82.29&81.08&81.35&75.95&15.03&31.76\\
        &MP&\usym{2718}&89.58&87.60&86.87&81.42&6.27&16.53&83.59&82.64&82.82&77.76&14.85&29.65\\ 
        &CAP&\usym{2714}&82.29&45.19&42.81&67.29&6.70&23.83&82.81&81.18&81.82&76.67&14.96&31.13\\
        &CCPP&\usym{2714}&88.54&78.49&80.48&79.40&6.21&17.57&84.64&83.86&83.92&79.20&13.26&27.08\\ 
        &SC-DVPP-Ser&\usym{2714}&90.10&84.99&86.92&81.85&6.27&17.38&86.20&85.41&85.85&81.32&11.84&25.44\\  
        &SC-DVPP-C-Ser&\usym{2714}&\textbf{91.15}&\textbf{87.92}&\textbf{87.98}&\textbf{84.25}&6.48&\textbf{15.24}&\textbf{86.72}&\textbf{86.09}&\textbf{86.54}&\textbf{82.04}&\textbf{11.73}&\textbf{25.06}\\
        \bottomrule
    \end{tabular}}  
\end{table*}


\textbf{Results on APTOS2019.} Table~\ref{tab:7}(Left) provides the imbalanced DR grading results of two DVPP implementations and state-of-the-art pooling methods on the APTOS2019. Note that we mainly apply bAcc, mF1, and kappa to assess classification performance due to the imbalanced data distribution. SC-DVPP-C-DVPP generally performs better than other advanced pooling methods regarding DR grading and confidence calibration performance. For example, compared to UPool and MP, SC-DVPP-C obtains above absolute 0.83\%  and 1.6\%  gains of bAcc and kappa, decreasing ECE value by 2.31\%. It is worth noting that SC-DVPP-C outperforms CAP and CCPP by absolute over \textbf{4.63\%} in bAcc, mF1, and kappa, while obtaining smaller values of ECE and BS.

\textbf{Results on NIH-CHR-LT.} Table~\ref{tab:7}(Right) offers the long-tailed medical image classification and confidence calibration comparisons of DVVP and other advanced pooling methods on the NIH-CHR-LT dataset. We also observe that our DVPP obtains better performance than these advanced pooling methods. Our SC-DVPP obtains 2.57\% and \textbf{7.21\%} gains of accuracy and kappa through comparisons to LIP and AdaPool, and it at least reduces \textbf{13.81\%} of ECE and BS.

\begin{table*}
    \centering
    \caption{Performance comparisons of different pooling methods in terms of medical image classification and confidence calibration performance on APTOS2019 and NIH-CXR-LT.}
    \label{tab:7}
  \resizebox{1\columnwidth}{!}{
    \begin{tabular}{c|cccccc|cccccc} 
        \toprule
         \multirow{2}{*}{Method}&\multicolumn{6}{c|}{APTOS2019}&\multicolumn{6}{c}{NIH-CXR-LT}\\
         &ACC$\uparrow$ & bAcc$\uparrow$ & mF1$\uparrow$&kappa$\uparrow$& ECE$\downarrow$ & BS$\downarrow$ &ACC$\uparrow$ & bAcc$\uparrow$ & mF1$\uparrow$&kappa$\uparrow$& ECE$\downarrow$ & BS$\downarrow$  \\
        \midrule
        GAP&86.95&70.35&72.35&79.39&10.42&23.85&50.61&12.26&11.01&12.05&23.88&74.23\\
        GMP&86.77&67.76&70.80&78.93&10.35&23.71&53.96&12.16&11.81&14.09&25.85&71.37\\
        SPP&86.77&71.18&72.92&79.16&10.40&23.72&49.52&12.01&9.96&10.09&34.33&82.33\\
        LIP&86.77&69.62&72.32&78.96&10.16&23.61&48.42&12.84&10.43&11.55&21.53&74.52\\
        S3Pool&86.21&69.03&70.29&78.34&\textbf{3.19}&20.85&53.52&11.56&10.43&13.25&19.00&69.03\\
        AdaPool&86.95&69.39&71.50&79.37&11.20&24.39&50.79&12.23&10.81&11.57&33.39&80.62\\
        UPool&86.95&69.40&72.56&79.36&4.70&20.78&50.98&12.28&10.78&12.05&25.97&74.32\\
        MP&86.95&71.31&73.21&79.41&9.79&23.30&51.60&12.11&11.36&12.40&31.13&78.19\\ 
        CAP&83.27&62.13&62.90&73.55&10.34&25.72&53.25&9.56&9.39&12.09&19.68&69.46\\
        CCPP&84.19&66.80&68.55&75.09&8.05&24.73&53.26&9.48&9.94&9.64&28.18&74.53\\

        SC-DVPP-Ser&86.95&71.03&72.66&79.49&6.66&21.61&\textbf{58.91}&10.31&10.15&12.51&9.24&\textbf{59.22}\\ 
        SC-DVPP-C-Ser&\textbf{88.24}&\textbf{71.43}&\textbf{74.07}&\textbf{81.45}&5.92&\textbf{20.46}&58.00&\textbf{12.29}&10.89&\textbf{14.14}&\textbf{5.20}&\textbf{60.81}\\
        \bottomrule
    \end{tabular}}
\end{table*}

\subsubsection{Validation on 3D Medical Image Datasets} Table~\ref{tab:9} presents 3D medical image classification and confidence calibration of our DVPP and other pooling methods on OASIS and ABIDE-I datasets. Here, we use the pre-trained 3D ResNet18 as the backbone network to investigate the effectiveness of our DVPP. For the OASIS dataset, the input volume is of size $169\times208\times179$. The initial learning rate, training epochs, and batch size are set to 0.001, 60, and 12 following \cite{wen2020convolutional}, respectively. For the ABIDE-I dataset, the input volume is of size $61\times73\times61$. The initial learning rate, training epochs, and batch size are set to 0.001, 60, and 32, respectively. The experimental results show that the two DVPP implementations generally outperform competitive pooling methods in terms of classification and confidence calibration performance, further validating the effectiveness and generalization capability of DVPP.

\begin{table*}
    \centering
    \caption{Performance comparisons of different pooling methods in terms of medical image classification and confidence calibration on OASIS and ABIDE-I. }
    \label{tab:9}
  \resizebox{0.98\columnwidth}{!}{
    \begin{tabular}{c|cccccc|cccccc} 
        \toprule
         \multirow{2}{*}{Method}&\multicolumn{6}{c|}{OASIS}&\multicolumn{6}{c}{ABIDE-I}\\
         &ACC$\uparrow$ & bAcc$\uparrow$ & mF1$\uparrow$&kappa$\uparrow$& ECE$\downarrow$ & BS$\downarrow$ &ACC$\uparrow$ & bAcc$\uparrow$ & mF1$\uparrow$&kappa$\uparrow$& ECE$\downarrow$ & BS$\downarrow$  \\
        \midrule
      
        GAP         &75.00&70.00&69.75&43.75&\textbf{14.97}&41.33&60.94&58.64&58.59&18.20&26.86&66.70\\
        GMP         &70.83&65.00&63.08&33.33&28.99&58.28&57.03&51.93&40.82&14.24&12.05&51.52\\
        SPP         &66.67&61.43&59.66&25.00&37.00&66.17&58.59&58.35&58.29&16.62&\textbf{7.84}&50.51\\
      
        LIP         &75.00&74.29&74.29&48.57&23.46&47.54&60.94&58.68&57.94&\textbf{18.80}&28.80&64.02\\
        S3Pool      &62.50&56.43&52.53&14.29&38.63&74.92&57.81&54.88&52.72&10.26&13.40&52.60\\
   
        UPool       &70.83&66.43&66.06&35.38&25.90&48.91&59.38&57.67&57.33&15.73&15.35&54.93\\
        MP          &79.17&79.29&78.84&57.75&24.82&43.27&\textbf{61.72}&58.06&54.50&17.21&10.68&49.84\\
        CCPP        &66.67&61.43&59.66&25.00&19.24&48.31&58.34&58.12&58.03&15.22&10.35&51.34\\ 
       
        SC-DVPP-Ser   &83.33&82.86&82.86&65.71&19.13&33.33&\textbf{61.72}&\textbf{59.23}&57.15&17.52&8.98&\textbf{49.59}\\  
        SC-DVPP-C-Ser &\textbf{87.50}&\textbf{86.43}&\textbf{86.93}&\textbf{73.91}&25.80&\textbf{33.07}&60.16&58.90&\textbf{58.83}&18.07&8.63&49.84\\
        \bottomrule
    \end{tabular}}
\end{table*}

\subsection{Performance Comparisons with Advanced Calibration Methods} 
Table~\ref{tab:10} provides the imbalanced image classification and confidence calibration results of calibration methods, pooling methods, and our DVPP on ISIC2018 and APTOS2019. 
It can be observed that our method performs better than comparable pooling and calibration methods in all classification evaluation and confidence calibration measures. Interestingly, several pooling methods not only achieve better classification results but also show competitive confidence calibration performance compared to advanced calibration methods. As we previously suggested, from the perspective of representation learning, all methods aim to guide DNNs to emphasize informative features and suppress redundant ones to achieve promising classification and confidence calibration. For example, some spatial pooling methods can aggregate significant spatial features, including rich salient features, thereby improving both classification and confidence calibration.
In particular, DVPP extracts both multi-scale spatial features and multi-scale pixel-wise features, incorporating both salient and subtle features. This explains why it achieves better classification and confidence calibration performance than other methods, which is consistent with our expectations.

\begin{table*}[t]
  \caption{Performance comparison of our DVPP, competitive pooling methods, and advanced calibration methods in terms of classification and confidence calibration on two imbalanced medical image datasets: ISIC2018 and APTOS2019.}
   \label{tab:10}
    \resizebox{0.98\columnwidth}{!}{
  \centering
  \begin{tabular}{c|c|cccc|cccc}
    \hline 
     \multirow{2}{*}{Category}& \multirow{2}{*}{Method}& \multicolumn{4}{c|}{ISIC2018} & \multicolumn{4}{c}{APTOS2019}\\
     &&mF1$\uparrow$&bAcc$\uparrow$&ECE$\downarrow$&BS$\downarrow$&mF1$\uparrow$&bAcc$\uparrow$&ECE$\downarrow$&BS$\downarrow$ \\
    \hline 
      \multirow{5}{*}{Calibration} & CE  &54.41&59.70&13.66&31.21& 70.35&72.35&10.42&23.85\\
   & Temp. \cite{guo2017calibration} &68.11&63.85&10.99&31.91&71.37&68.39&7.93&21.64\\
   & FCL~\cite{lin2017focal}    &64.31&64.43&11.14&32.64&66.42&63.94&12.76&23.20\\
   & LDAM~\cite{cao2019learning}  &52.40&59.95&13.31&46.68&70.28&68.42&10.84&24.68\\ 
   & Mixup~\cite{zhang2018mixup} &65.16&65.48&9.53&28.84&68.31&65.06&6.84&21.15\\ \hline
   \multirow{6}{*}{Pooling}& SPP &75.55&78.49&9.47&23.92 &71.18&72.92&10.40&23.72\\
   & LIP&74.30&73.63&9.48&23.92&69.62&72.32&10.16&23.61\\
   & MP &77.86&80.57&9.14&23.52&70.60&74.01&10.65&23.11\\
   & CCPP&77.51&73.41&10.07&24.66 &66.80&68.55&8.05&24.73\\ 
    & SC-DVPP-Ser&81.33&78.59&9.83&21.63&\textbf{72.66}&71.03&6.66&21.61\\
   & SC-DVPP-C-Ser&\textbf{82.91}&\textbf{83.43}&\textbf{8.70}&\textbf{21.52}&71.43&\textbf{74.07}&\textbf{5.92}&\textbf{20.46}\\
     \hline
  \end{tabular}}
 
\end{table*}

In general, we argue the advantages of our DVPP over existing SOTA pooling methods and calibration methods as follows: 
\begin {itemize}
\item[1.] As previously introduced, previous SP and CCP methods have a high possibility of losing subtle feature representations or salient feature representations, owing to the lack of spatial features or pixel-wise features. On the contrary, our proposed DVPP is the first pooling method to obtain both multi-scale spatial and multi-scale pixel-wise features at the same time, involving enough subtle and salient feature representations that are beneficial for medical classification results and confidence calibration boosting.

\item[2.] From the aspect of representation learning, most existing calibration methods focus on the learnable weight parameter optimization for generating informative feature maps; thereby, widely used SP methods can aggregate useful spatial features but inevitably ignore the relative importance of pixel-wise features in confidence calibration improvement. Differently, our method aggregates multi-scale dual-view features encompassing both spatial features and pixel-wise features simultaneously, which provides sufficient yet valuable features for classification and confidence calibration boost. \textbf{Moreover, our method provides a new perspective to enhance classification and confidence calibration improvement from the aspect of pooling, which may prompt the development of this field.}

\item[3.] We construct efficient implementations of DVPP from a dual-axis perspective, conducing to mine the potential of spatial features and pixel-wise features effectively.

\end {itemize}

\subsection{Visual Analysis and Explanation}

\subsubsection{Visualization of Multi-Scale Dual-View Features in SC-DVPP}
To investigate the inherent behavior of our DVPP, we visualize its feature distributions on the testing images of ISIC2018 and BTM. Here, we take the SC-DVPP-C-Ser as the representative implementation of DVPP, including multi-scale dual-view features and multi-scale pixel-wise features.

\begin{figure*}
\centering
\centerline{\includegraphics[width=1.0\linewidth,height=8.5cm]{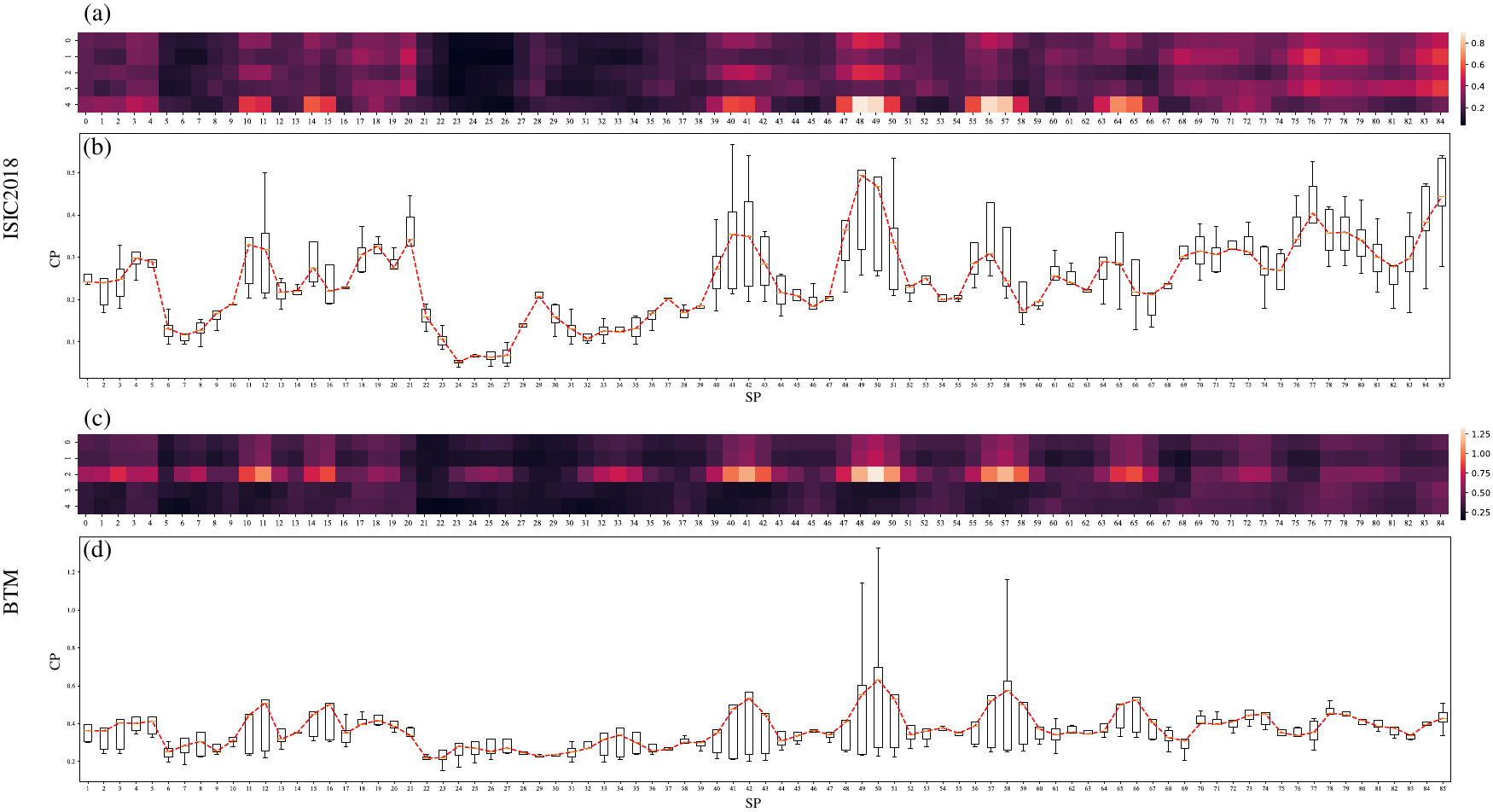}}
  \caption{The multi-scale dual-view feature maps and multi-scale dual-view feature statistics of SC-DVPP in SC-DVPP-C. The datasets are ISIC2018 and BTM.}
	\label{fig6}
\end{figure*}

\textbf{Multi-scale dual-view feature visualization.}
Fig.~\ref{fig6}(a)-(b) presents multi-scale dual-view feature maps and multi-scale dual-view feature statistics of SC-DVPP on a representative test image from the ISIC2018 dataset. We observe as follows: 1) Multi-scale dual-view feature values are different from each other, indicating the significance of them with varying levels. 2) We see the noticeable multi-scale dual-view feature differences among multi-scales, showing the spatial and pixel-wise features at different scales, impacting classification and confidence calibration results at different levels. Visualizations on the representative MRI image also show a similar conclusion, as shown in Fig.~\ref{fig6}(c)-(d).

\begin{figure*}
\centering
\centerline{\includegraphics[width=0.95\linewidth]{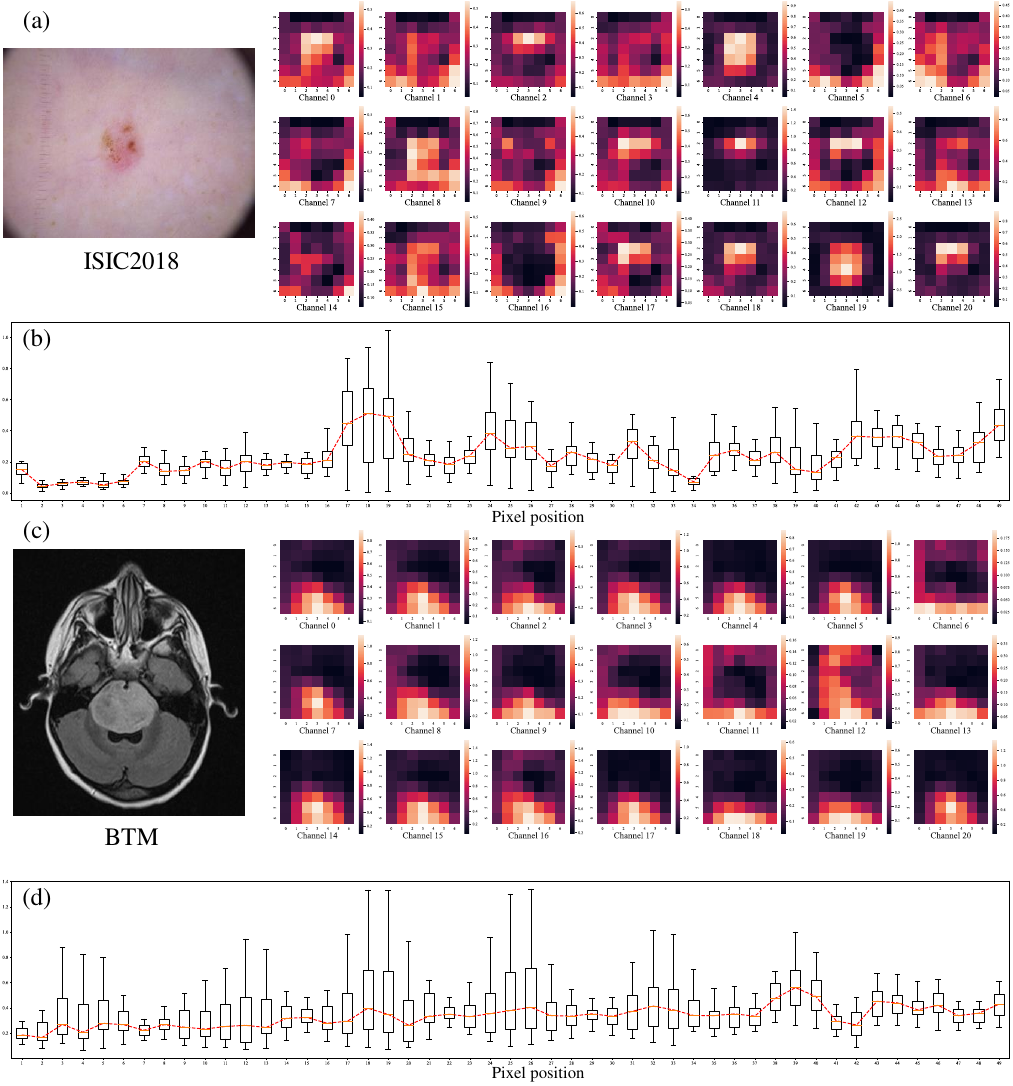}}
  \caption{(a)-(b): multi-scale pixel-wise feature map and pixel-wise feature value distribution visualizations of CCP operator in SC-DVPP-C-Ser on the ISIC2018. (c)-(d): multi-scale pixel-wise feature map and pixel-wise feature value distribution visualizations of CCP operator in SC-DVPP-C-Ser on the ISIC2018.}
	\label{fig7}
\end{figure*}

\textbf{Multi-scale pixel-wise feature visualization.} Fig.~\ref{fig7}(a) provides the multi-scale pixel-wise feature map visualizations of CCP operator in SC-DVPP-C-Ser on the ISIC2018, and we see that visual differences among these pixel-wise feature maps, indicating the significances of them are different to each other. Moreover, Fig.~\ref{fig7}(b) offers the pixel-wise feature value distribution visualizations along the pixel position axis in all multi-scale pixel-wise feature maps, highlighting that their roles differ at various pixel positions. Additionally, we also obtain similar conclusions on the BTM dataset, as offered in Fig.~\ref{fig7}(c)-(d).

\subsubsection{Feature Representation Visualization}
\begin{figure*}
\centering
\centerline{\includegraphics[width=0.95\linewidth,height=5cm]{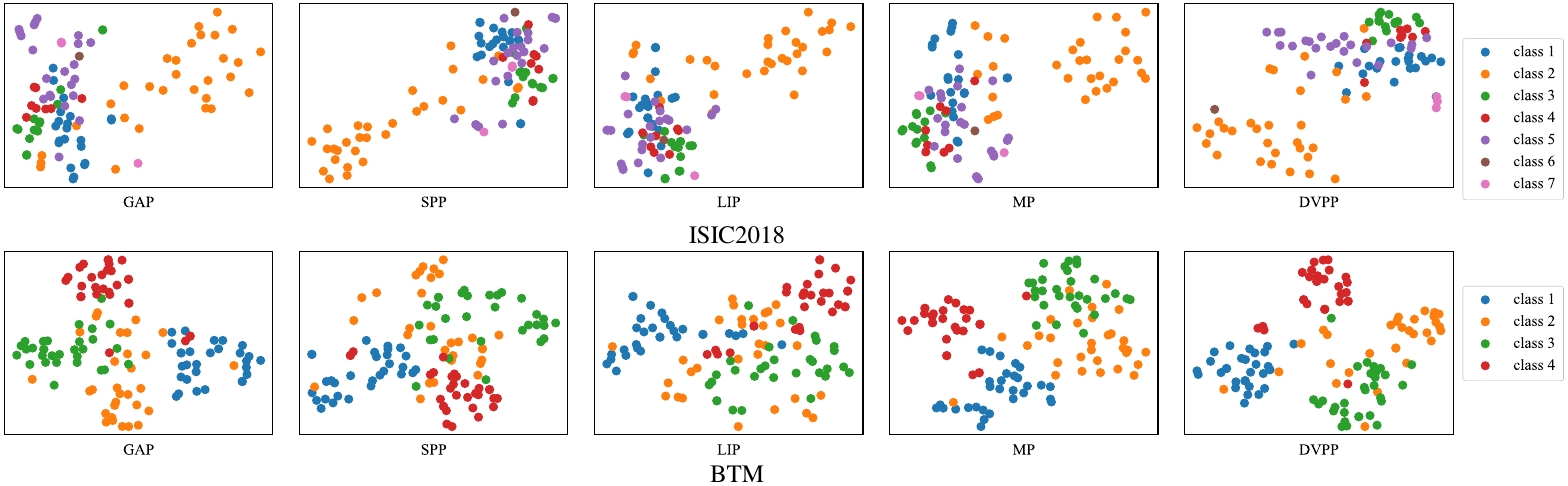}}
  \caption{t-SNE visualizations of features aggregated by our DVPP and other competitive pooling methods on ISIC2018 and BTM.}
\label{fig8}
\end{figure*}

To investigate the effects of our DVPP more deeply, Fig.~\ref{fig8} presents t-SNE visualizations of features obtained by GAP, SPP, LIP, MP, and our DVPP on ISIC2018 and BTM. The class clustering distributions of our DVPP differ from those of comparable pooling methods, highlighting the marked representation differences between the extracted dual-view features and spatial features. Moreover, we observe that the features extracted by our DVPP are more compact within individual classes and more separated among different classes than those extracted by other pooling methods. This demonstrates that our DVPP significantly impacts medical image classification and confidence calibration results.

\section{Conclusion and Future Work}
\label{sec:con}

This paper presents a novel dual-view framework to systematically analyze the differences between SP and CCP operators. We suggest that these two pooling operators have unique impacts on classification and confidence calibration results. Under this framework, we develop a dual-view pyramid pooling (DVPP) to mine the potential of multi-scale dual-view features from a dual-axis perspective, plugged at the end of DNNs, to improve medical image classification and confidence calibration performance. Experimental results show that our method outperforms state-of-the-art methods, agreeing with our expectations. Moreover, the theoretical analysis of the proposed is insufficient and will be addressed in the future. We also plan to modify DVPP, which is applicable to other downstream learning tasks, such as image segmentation and objection detection.

\backmatter
\bmhead{Data Availability}
Experimental  results of Tables~\ref{tab:1} -\ref{tab:10} and Figures~\ref{fig6},\ref{fig7},\ref{fig8} based on six publicly available 2D/3D medical image datasets \cite{tschandl2018ham10000,BTM,APTOS,holste2022long,wen2020convolutional,di2014autism}.

\bmhead{Acknowledgments}
This work was partly supported by the Key Research and Development Project of Guangdong Province (No. 2021B0101310002), National Science Foundation of China (No. 62272449), the Shenzhen Basic Research Fund (No. KQTD20200820113106007), and Shenzhen Key Laboratory of Intelligent Bioinformatics(No.ZDSYS20220422103800001). We would also like to thank the funding support by the Youth Innovation Promotion Association(No.Y2021101), CAS to Yanjie Wei.
\bibliographystyle{bst/sn-chicago}
\bibliography{sn-bibliography}

\end{document}